\documentclass[11pt]{article}

\usepackage[T1]{fontenc}
\usepackage{graphicx}
\usepackage{chemformula}
\usepackage{microtype}
\usepackage{amsmath,amssymb}
\usepackage{booktabs}
\usepackage{siunitx}
\usepackage{hyperref}
\usepackage{tabularx}
\usepackage{float}
\usepackage{geometry}
\usepackage{caption}
\usepackage{subcaption}
\usepackage{parskip}
\usepackage{placeins}
\usepackage{cleveref}
\usepackage{longtable}
\usepackage{multirow}
\usepackage{multicol}
\usepackage{pdflscape}
\usepackage{rotating}
\usepackage{seqsplit}
\usepackage{cite}

\hypersetup{
    colorlinks=true,
    linkcolor=blue,
    citecolor=blue,
    urlcolor=blue
}

\title{\bfseries Data Leakage and Redundancy in the \textsc{LIT-PCBA} Benchmark}
\author{
  Amber Huang, Ian Scott Knight, Slava Naprienko\\
  \small \texttt{\{amber, ian, slava\}@sievestack.com} \\
  \small SieveStack, Inc.
}
\date{}

\begin{document}
\maketitle

\begin{abstract}
\textsc{LIT-PCBA} is widely used to benchmark virtual screening models, but our audit reveals it is fundamentally compromised. We find extensive data leakage and molecular redundancy across its data splits, including 2D-identical ligands within splits, pervasive analog overlap, and low‑diversity query sets. In \textsc{ALDH1} alone, for example, 323 active training–validation analog pairs occur at ECFP4 Tanimoto $\geq 0.6$; across all targets, 2{,}491 2D-identical inactives appear in both training and validation, with exceptionally few actives that are 2D‑identical to these inactives. Such overlap allows models to succeed via scaffold memorization rather than generalization, inflating enrichment factors and AUROC scores.

These flaws are not incidental: they are so severe that a trivial memorization-based baseline with no learnable parameters exploits these artifacts to match or exceed the reported performance of state‑of‑the‑art deep learning and 3D‑similarity models.

These data integrity failures undermine nearly all published \textsc{LIT-PCBA} results. Even models evaluated in ``zero‑shot'' mode are affected by analog leakage into the query set, undermining claims of generalization. In its current form, the benchmark does not measure the intended ability to recover novel chemotypes and should not be treated as evidence of methodological progress. 

All scripts necessary to reproduce our audit and the baseline implementation are available at: \url{https://github.com/sievestack/LIT-PCBA-audit}.

\end{abstract}

\section{Introduction}

The landscape of computational drug discovery has been transformed by the development and widespread adoption of robust benchmarking datasets~\cite{lagarde2015benchmarking, xia2015benchmarking, irwin2008community}. Among these, the \textsc{LIT-PCBA} benchmark~\cite{tran2020lit} was recently introduced as a response to well-documented limitations in earlier datasets like \textsc{DUD-E} and \textsc{MUV}, which have been criticized for analog bias, decoy artifacts, and inflated performance due to confounding structural patterns~\cite{chen2019hidden, mysinger2012directory, rohrer2009maximum, scior2012recognizing}. To address these issues, \textsc{LIT-PCBA} combined thousands of experimentally validated actives and hundreds of thousands of inactives from PubChem bioassays with strategies like asymmetric validation embedding (AVE)~\cite{rohrer2009maximum} to reduce spurious correlations and improve the rigor of model evaluation. A third component — the query set — comprises ligands co-crystallized with each protein target, used as fixed references to evaluate model performance on unseen compounds. While limitations remain~\cite{scior2012recognizing}, \textsc{LIT-PCBA} represents a thoughtful attempt to raise the standard for both ligand- and structure-based virtual screening.

Since its release in 2020~\cite{tran2020lit}, \textsc{LIT-PCBA} has rapidly emerged as a \textit{de facto} gold standard for ML-based virtual screening, frequently cited in reviews and methods papers~\cite{ghislat2021recent, kimber2021deep, murugan2022artificial}. A wave of new models — spanning protein language models, graph neural networks, 3D similarity approaches, and contrastive learning — report state-of-the-art 1\% enrichment factor ($\mathrm{EF}_{1\%}$) scores on \textsc{LIT-PCBA}~\cite{sim2025recent, gorgulla2023recent}. Prominent pipelines such as \textbf{SPRINT}~\cite{mcnutt2025scaling}, \textbf{DrugCLIP}~\cite{gao2023drugclip}, \textbf{DENVIS}~\cite{krasoulis2022denvis}, and \textbf{CHEESE}~\cite{lvzivcavr2024cheese} claim superior $\mathrm{EF}_{1\%}$ values on \textsc{LIT-PCBA} that surpass those achieved by traditional docking and 2D similarity methods. Similarly, consensus and meta-learning-based affinity predictors like \textbf{MILCDock}~\cite{morris2022milcdock} and deep learning-based docking frameworks such as \textbf{GNINA}~\cite{gnina2021} and improved \textbf{DeepDTA}~\cite{li2024deepdta} have been extensively benchmarked  on \textsc{LIT-PCBA}.

In this work, we present a rigorous audit of data integrity failures in \textsc{LIT-PCBA}. Despite its widespread adoption, our audit reveals severe data leakage, molecular redundancy, and analog bias — issues that artificially inflate model performance and undermine claims of generalizability. We show that these flaws systematically distort performance metrics — particularly $\mathrm{EF}_{1\%}$ and AUROC — and raise serious concerns about the use of \textsc{LIT-PCBA} as a benchmark for virtual screening. Our findings parallel the conclusions of several recent audits and meta-analyses uncovering fundamental flaws in benchmarks like \textsc{DUDE-Z}~\cite{chen2019hidden, janet2023latentbias, kanakala2023latent}, which suggest that many putative state-of-the-art results achieved on such benchmarks may reflect covert memorization rather than any meaningful advances in molecular modeling. 

The consequences of these flaws are far-reaching, directly impacting the interpretation of results for nearly every modern virtual screening method. High-performing deep neural network encoders, including prominent models like SPRINT, DrugCLIP, and BIND, have reported impressive enrichment factors on \textsc{LIT-PCBA}. However, our analysis reveals that these high scores can be largely attributed to the pervasive data leakage and redundancy. As we demonstrate, a trivial memorization-based model — with no learned chemical or biophysical insights — can match or even surpass the performance of these sophisticated architectures simply by exploiting the overlap between training, validation, and query sets.

Therefore, any claim of state-of-the-art performance on \textsc{LIT-PCBA} must be treated with substantial skepticism. Even models evaluated in a ``zero-shot'' setting are not immune, as analog leakage ensures that the test compounds in the query set often closely resemble molecules seen during pre-training. In its current state, the benchmark fails to assess the ability of a model to generalize to novel chemotypes — the central goal of real-world virtual screening. As a result, previously reported performance metrics on \textsc{LIT-PCBA} can no longer be regarded as evidence of methodological progress, as they are likely inflated by exploitation of benchmark-specific artifacts.

To help advance the field, we introduce an analog-aware auditing framework and suggest new protocols for future benchmark construction, responding to long-standing calls for more rigorous, reliability-centered evaluation that will enable robust, generalizable ML for molecular discovery~\cite{lagarde2015benchmarking, irwin2008community, schneider2010virtual}.

\textbf{Acknowledgments}. We thank Bogdan Zavyalov for helpful remarks, and Miroslav L\v{z}i\v{c}a\v{r} for clarifying the CHEESE evaluation protocol, which enabled an apples-to-apples comparison.

\section{\textsc{LIT-PCBA} Benchmark}
The \textsc{LIT-PCBA} (Literature-derived PubChem BioAssay) benchmark~\cite{tran2020lit} comprises 15 protein targets, each with at least one co-crystal structure and curated bioactive/inactive compounds from PubChem. Structurally, it follows the paradigm of earlier benchmarks such as DUD-E and DUDE-Z~\cite{mysinger2012directory, stein2021property}, which pair crystallographic references with ligand-ranking tasks. For each target, \textsc{LIT-PCBA} provides three components: a set of query ligands from co-crystal structures, a training set for model fitting, and a validation set for evaluation. The query ligands are intended to represent ``unseen'' actives against which generalization is measured, and training/validation compounds are partitioned using the asymmetric validation embedding (AVE) protocol~\cite{wallach2018most}. Target statistics are in \Cref{tab:lit_pcba_data}.

\begin{table}[H]
\setlength{\tabcolsep}{2pt}
\centering
\begin{tabular}{
  l
  c
  *{6}{S[table-format=6.0,table-number-alignment=center]}
}
\toprule
\textbf{Target} & \textbf{PDBs} & \textbf{Act.} & \textbf{Act. (T)} & \textbf{Act. (V)} & \textbf{Inact.} & \textbf{Inact. (T)} & \textbf{Inact. (V)} \\
\midrule
ADRB2     & 8  & 17    & 13   & 4    & 311748 & 233957 & 77791 \\
ALDH1     & 8  & 5363  & 4020 & 1343 & 101874 & 76577  & 25297 \\
ESR1 (ago) & 15 & 13    & 10   & 3    & 4378   & 3470   & 908 \\
ESR1 (ant) & 15 & 88    & 63   & 25   & 3820   & 3026   & 794 \\
FEN1      & 1  & 360   & 269  & 91   & 350718 & 263771 & 86947 \\
GBA       & 6  & 163   & 122  & 41   & 291241 & 219042 & 72199 \\
IDH1      & 14 & 39    & 30   & 9    & 358757 & 269664 & 89093 \\
KAT2A     & 3  & 194   & 146  & 48   & 342729 & 258067 & 84662 \\
MAPK1     & 15 & 308   & 231  & 77   & 61567  & 46317  & 15250 \\
MTORC1    & 11 & 97    & 73   & 24   & 32972  & 24729  & 8243 \\
OPRK1     & 1  & 24    & 18   & 6    & 269475 & 202166 & 67309 \\
PKM2      & 9  & 546   & 410  & 136  & 244679 & 183672 & 61007 \\
PPARG     & 15 & 24    & 18   & 6    & 4071   & 3210   & 861 \\
TP53      & 6  & 64    & 48   & 16   & 3345   & 2550   & 795 \\
VDR       & 2  & 655   & 490  & 165  & 262648 & 197557 & 65091 \\
\bottomrule
\end{tabular}
\caption{Per-target summary for \textsc{LIT-PCBA}. Columns (left to right): \textit{Target}; \textit{PDBs} (number of query structures); \textit{Act.} (total actives); \textit{Act. (T)} (training actives); \textit{Act. (V)} (validation actives); \textit{Inact.} (total inactives); \textit{Inact. (T)} (training inactives); \textit{Inact. (V)} (validation inactives).}
\label{tab:lit_pcba_data}
\end{table}

There are two primary ways to use datasets like DUDE-Z or \textsc{LIT-PCBA} for ligand-based methods. In both cases, the first step is to extract one or more ligands are from co-crystallized PDB structures; these are the \emph{queries}, which serve as the fixed inputs for similarity-based virtual screening. See \cite{jiang2021comprehensive, wang2024conformational, krasoulis2022denvis, yang2022protein, yang2022delta, lvzivcavr2024cheese} for examples following this exact protocol.

\begin{enumerate}
    \item \textbf{Model training and selection.} If \textsc{LIT-PCBA} is used for training, the queries are used to score \emph{all} molecules in the training set (actives and inactives). Ranking metrics (e.g., nEF$_{1\%}$, average precision, ROC AUC, BEDROC) are computed and used to guide model development, including hyperparameter tuning and training procedure adjustments. Best practice is to further split the training set into internal training and validation subsets for this purpose, treating the \textsc{LIT-PCBA} ``validation'' split as a held-out \emph{test} set in standard ML terminology.
    \item \textbf{Final evaluation.} Once tuning is complete, the trained model and fixed queries are used to score the test set (the \textsc{LIT-PCBA} ``validation'' split) \emph{without further modification or retraining}. Final metrics are reported on this held-out set to avoid post-hoc overfitting.
\end{enumerate}

Two evaluation strategies are common. 
\begin{enumerate}
    \item \textbf{Single-ligand evaluation} ranks the library using one query at a time; when multiple PDB-derived queries are available, it is standard to report the mean performance across individual query results. Of course, reporting per-query metrics separately is more transparent and scientifically precise.
    \item \textbf{Multi-ligand evaluation} ranks molecules using all queries simultaneously. The original \textsc{LIT-PCBA} paper adopts simple \emph{max-pooling} over queries: for each molecule, compute its similarity to every query and rank by the maximum. While alternative fusion strategies exist, max-pooling remains the most common and transparent method for aggregating query signals.
\end{enumerate} 

For most targets, multiple query ligands are available from distinct PDB entries. Several entries, however, correspond to the \emph{same} protein–ligand complex, so multiple structures do not necessarily yield distinct query ligands. This can help structure-based methods (sampling conformational variability), but for ligand-based methods — which use only ligand identity — it introduces repetitions that must be handled carefully.

To ensure consistent ligand identities, we manually curated each query ligand from RCSB PDB and derived canonical SMILES with RDKit~\cite{rdkit}. The resulting isomeric SMILES and the PDB–ligand mapping appear in \Cref{sec:appendix} (\Cref{tab:receptor_pdb_ligands}); full curation details are provided there.

\section{Memorization-Based Model Baseline}\label{sec:cheating}
The central finding of this audit is striking: a trivial, stereo‑agnostic memorization baseline with no learnable parameters matches the median EF$_{1\%}$ of state‑of‑the‑art 3D encoders on \textsc{LIT‑PCBA}.
This decisive result is independent of any debate over how to define ``data leakage''. \emph{Before any curation}, we evaluated the dataset exactly as released, using the provided SMILES, train/validation/query splits, and activity labels. That this trivial baseline succeeds on the unaltered dataset shows the flaws are inherent to the benchmark, not introduced by our later audit, where stereochemistry was removed only to expose potential sources of leakage.

\subsection{Comparison to \textsc{CHEESE} 3D Encoders}

The stereo‑agnostic memorization baseline is built from 4096‑bit ECFP4 fingerprints and uses only the training actives as a reference set. No learning, physics, or modeling is involved; each test molecule is scored solely by its maximum fingerprint similarity to a training active. Despite its triviality, the baseline reaches a median raw EF${1\%}$ of 4.15 on the validation set (see \Cref{tab:ef1_multi}, \textbf{EF$_{1\%}$}, ``Mem''). 
This score is on par with the state‑of‑the‑art median raw EF${1\%}$ reported for the \textsc{CHEESE} 3D encoders (EspSim and ShapeSim) under the same multi-ligand protocol (Fig.~7 of \cite{lvzivcavr2024cheese}).
The implication is unavoidable: either \textsc{LIT‑PCBA}’s performance ceiling is dominated by leakage that rewards memorization, or the 3D encoders offer no measurable advantage over a trivial 2D baseline.

\begin{table}[ht]
\footnotesize
\setlength{\tabcolsep}{4pt}
\centering
\begin{tabular}{
  l
  *{6}{S[table-format=2.2]}
  |
  *{6}{S[table-format=1.3]} }
\toprule
\multirow{2}{*}{\textbf{Target}} &
\multicolumn{6}{c|}{\textbf{EF\textsubscript{1\%}}} &
\multicolumn{6}{c}{\textbf{nEF\textsubscript{1\%}}}\\
\cmidrule(lr){2-7}\cmidrule(lr){8-13}
& \multicolumn{1}{c}{Mem} & \multicolumn{1}{c}{CE-E} & 
  \multicolumn{1}{c}{CE-S} & \multicolumn{1}{c}{ECFP4} & 
  \multicolumn{1}{c}{ROCS} & \multicolumn{1}{c|}{SFX} &
  \multicolumn{1}{c}{Mem} & \multicolumn{1}{c}{CE-E} & 
  \multicolumn{1}{c}{CE-S} & \multicolumn{1}{c}{ECFP4} & 
  \multicolumn{1}{c}{ROCS} & \multicolumn{1}{c}{SFX}\\
\midrule
ADRB2   & 0.00 & 0.00 & 0.00 & 0.00 & 0.00 & 0.00 & 0.000 & 0.000 & 0.000 & 0.000 & 0.000 & 0.000\\
ALDH1   & 5.00 & 5.13 & 4.24 & 2.68 & 1.64 & 0.82 & 0.252 & 0.259 & 0.214 & 0.135 & 0.083 & 0.041\\
ESR1 (ago) & 0.00 & 0.00 & 0.00 & 0.00 & 0.00 & 0.00 & 0.000 & 0.000 & 0.000 & 0.000 & 0.000 & 0.000\\
ESR1 (ant) & 4.10 & 3.64 & 7.28 & 0.00 & 4.00 & 4.00 & 0.125 & 0.111 & 0.222 & 0.000 & 0.122 & 0.122\\
FEN1    & 5.50 & 13.18 & 9.88 & 1.09 & 0.00 & 13.18 & 0.055 & 0.132 & 0.099 & 0.011 & 0.000 & 0.132\\
GBA     & 14.64 & 17.06 & 12.19 & 2.44 & 0.00 & 4.88 & 0.146 & 0.170 & 0.122 & 0.024 & 0.000 & 0.049\\
IDH1    & 11.11 & 0.00 & 22.20 & 0.00 & 0.00 & 0.00 & 0.111 & 0.000 & 0.222 & 0.000 & 0.000 & 0.000\\
KAT2A   & 4.17 & 8.32 & 2.08 & 0.00 & 0.00 & 2.08 & 0.042 & 0.083 & 0.021 & 0.000 & 0.000 & 0.021\\
MAPK1   & 7.81 & 7.76 & 6.46 & 1.30 & 2.60 & 1.30 & 0.078 & 0.077 & 0.064 & 0.013 & 0.026 & 0.013\\
MTORC1  & 4.15 & 4.15 & 4.15 & 0.00 & 0.00 & 4.17 & 0.042 & 0.042 & 0.042 & 0.000 & 0.000 & 0.042\\
OPRK1   & 0.00 & 0.00 & 0.00 & 16.67 & 0.00 & 0.00 & 0.000 & 0.000 & 0.000 & 0.167 & 0.000 & 0.000\\
PKM2    & 1.47 & 5.14 & 2.20 & 0.74 & 2.21 & 0.74 & 0.015 & 0.051 & 0.022 & 0.007 & 0.022 & 0.007\\
PPARG   & 0.00 & 0.00 & 0.00 & 16.67 & 16.67 & 0.00 & 0.000 & 0.000 & 0.000 & 0.173 & 0.173 & 0.000\\
TP53    & 0.00 & 0.00 & 5.63 & 0.00 & 0.00 & 0.00 & 0.000 & 0.000 & 0.111 & 0.000 & 0.000 & 0.000\\
VDR     & 7.27 & 5.45 & 1.82 & 3.64 & 0.00 & 0.00 & 0.073 & 0.055 & 0.018 & 0.036 & 0.000 & 0.000\\
\midrule
Mean    & 4.35 & 4.66 & \textbf{5.21} & 3.02 & 1.81 & 2.08 & 0.063 & 0.065 & \textbf{0.077} & 0.038 & 0.028 & 0.028\\
Median  & \textbf{4.15} & \textbf{4.15} & \textbf{4.15} & 0.74 & 0.00 & 0.74 & 0.042 & \textbf{0.051} & 0.042 & 0.007 & 0.000 & 0.007\\
\bottomrule
\end{tabular}
\caption{``Multi-ligand'' experiment reproduced from Fig.\,7 of the \textsc{CHEESE} paper\,\cite{lvzivcavr2024cheese} with our memorization baseline included (``Mem''). ``CE-E'' and ``CE-S'' denote the EspSim and ShapeSim 3D encoders from the CHEESE framework, respectively; remaining methods are defined in the CHEESE paper~\cite{lvzivcavr2024cheese}. Left block: raw EF\textsubscript{1\%}. Right block: values normalized to each target’s theoretical maximum (nEF\textsubscript{1\%}). }
\label{tab:ef1_multi}
\end{table}

\subsubsection{\texorpdfstring{Limitations of Raw EF\textsubscript{1\%}}{Limitations of Raw EF1\%}}
Raw $\mathrm{EF}{1\%}$ values are scale‑dependent: the denominator changes with the number of screened molecules and with the active‑to‑inactive ratio, both of which vary widely across \textsc{LIT‑PCBA} targets. In general, reporting more statistics — and ideally releasing the full evaluation code and per‑target results — is the most transparent practice. At a minimum, both the mean and the median across targets should be reported, as relying on a single statistic allows even a single outlier — whether large or small, easy or hard — to distort the result. Moreover, each score should be normalized by that target’s theoretical maximum, yielding the dimensionless n$\mathrm{EF}{1\%}$. This triplet \{\textit{mean}, \textit{median}, \textit{n}$\mathrm{EF}_{1\%}$\} is not sufficient to characterize performance in every case, but it is necessary to distinguish genuine enrichment of actives from ostensible gains due solely to favorable class balance.

Applying these safeguards to the published results of the \textsc{CHEESE} 3D encoders makes the risks of relying on raw EF${1\%}$ explicit. Under the \emph{multi-ligand} protocol (\Cref{tab:ef1_multi}), the memorization baseline matches the CHEESE encoders on raw EF${1\%}$. After normalization, however, ShapeSim leads on the mean while EspSim leads on the median.

The effect is even sharper when re-evaluating the \emph{single‑ligand} protocol. In the original \textsc{CHEESE} paper, the authors report only the \emph{median} raw EF$_{1\%}$ — the one statistic that happens to favor their 3D encoder. As shown in \Cref{tab:ef1_single}, calculating the \emph{mean} directly from the raw scores listed in Fig.~6 of \cite{lvzivcavr2024cheese} reveals that their chosen ECFP4 baseline scores 2.49, surpassing EspSim and ShapeSim (2.18 each). Normalizing EF$_{1\%}$ widens the gap even more. (Note: the memorization baseline is undefined when there is a single reference ligand and is therefore omitted in \Cref{tab:ef1_single}.)

\begin{table}[ht]
\small
\setlength{\tabcolsep}{4pt}
\centering
\begin{tabularx}{\linewidth}{%
  l                                   
  *{5}{>{\centering\arraybackslash}X} 
  |                                   
  *{5}{>{\centering\arraybackslash}X}}
\toprule
\multirow{2}{*}{\textbf{Target}} &
\multicolumn{5}{c|}{\textbf{EF\textsubscript{1\%} (raw)}} &
\multicolumn{5}{c}{\textbf{nEF\textsubscript{1\%}}}\\
\cmidrule(lr){2-6}\cmidrule(lr){7-11}
& CE-E & CE-S & ECFP4 & ROCS & SFX
& CE-E & CE-S & ECFP4 & ROCS & SFX \\
\midrule
ADRB2     & 9.37 & 3.12 & 0.00 & 0.00 & 0.00 & 0.094 & 0.031 & 0.000 & 0.000 & 0.000 \\
ALDH1     & 1.20 & 1.18 & 1.58 & 1.08 & 1.25 & 0.060 & 0.059 & 0.080 & 0.054 & 0.063 \\
ESR1 (ago) & 2.02 & 8.10 & 0.00 & 0.00 & 0.00 & 0.020 & 0.080 & 0.000 & 0.000 & 0.000 \\
ESR1 (ant) & 1.21 & 0.49 & 2.67 & 1.07 & 1.60 & 0.037 & 0.015 & 0.081 & 0.033 & 0.049 \\
FEN1      & 2.20 & 2.20 & 1.09 & 0.00 & 3.26 & 0.022 & 0.022 & 0.011 & 0.000 & 0.033 \\
GBA       & 0.00 & 0.41 & 1.63 & 0.81 & 4.47 & 0.000 & 0.004 & 0.016 & 0.008 & 0.045 \\
IDH1      & 1.59 & 1.59 & 1.59 & 0.79 & 0.79 & 0.016 & 0.016 & 0.016 & 0.008 & 0.008 \\
KAT2A     & 1.39 & 3.47 & 0.69 & 0.69 & 4.17 & 0.014 & 0.035 & 0.007 & 0.007 & 0.042 \\
MAPK1     & 1.46 & 0.86 & 0.95 & 1.39 & 1.99 & 0.015 & 0.009 & 0.009 & 0.014 & 0.020 \\
MTORC1    & 0.00 & 3.77 & 0.00 & 0.00 & 1.52 & 0.000 & 0.038 & 0.000 & 0.000 & 0.015 \\
OPRK1     & 0.00 & 0.00 & 16.67 & 0.00 & 0.00 & 0.000 & 0.000 & 0.167 & 0.000 & 0.000 \\
PKM2      & 1.63 & 1.80 & 1.31 & 2.13 & 0.90 & 0.016 & 0.018 & 0.013 & 0.021 & 0.009 \\
PPARG     & 1.07 & 3.21 & 5.56 & 5.56 & 5.56 & 0.011 & 0.033 & 0.058 & 0.058 & 0.058 \\
TP53      & 2.82 & 0.00 & 0.00 & 0.88 & 0.00 & 0.056 & 0.000 & 0.000 & 0.017 & 0.000 \\
VDR       & 6.66 & 2.42 & 3.64 & 0.00 & 0.00 & 0.067 & 0.024 & 0.036 & 0.000 & 0.000 \\
\midrule
Median    & 1.46 & \textbf{1.80} & 1.31 & 0.79 & 1.25 & 0.016 & \textbf{0.022} & 0.013 & 0.008 & 0.015 \\
Mean      & 2.18 & 2.18 & \textbf{2.49} & 0.96 & 1.70 & 0.029 & 0.026 & \textbf{0.033} & 0.015 & 0.023 \\
\bottomrule
\end{tabularx}
\caption{Single-ligand experiment reproduced from Fig.\,6 of the \textsc{CHEESE} paper\,\cite{lvzivcavr2024cheese}. ``CE-E'' and ``CE-S'' denote the EspSim and ShapeSim 3D encoders from the CHEESE framework, respectively; remaining methods are defined in the CHEESE paper~\cite{lvzivcavr2024cheese}. Raw EF\textsubscript{1\%} scores are shown on the left; the right panel reports scores normalized to each target’s theoretical maximum (nEF\textsubscript{1\%}). Bold numbers denote the best method per row. Note: the memorization baseline is undefined when there is a single reference ligand and is therefore omitted in \Cref{tab:ef1_single}.}
\label{tab:ef1_single}
\end{table}

\subsubsection{Impact of Selective Metric Reporting}

Selective reporting of a single summary statistic or relying on unnormalized scores can invert the apparent ranking of methods, masking weaknesses and inflating strengths. As shown in this comparison with \textsc{CHEESE}, when both mean and median are shown side by side, with scores normalized to each target’s theoretical maximum, the supposed 3D advantage vanishes; which model ``wins'' depends entirely on which statistic the reader sees. Without full disclosure of mean, median, and n$\mathrm{EF}_{1\%}$ — and the code to reproduce them — claims of state-of-the-art performance cannot be taken at face value.

\subsection{Non-learning Baseline Variants}

We also evaluated several related \emph{non‑learning} baselines, all built from the same 4096‑bit ECFP4 fingerprints. These variants differ only in the choice of the reference set and in how the two sources of prior knowledge — training actives and PDB‑derived queries — are combined.  
\Cref{tab:maxpool_variants} summarizes their performance, showing that using training actives alone (i.e., our reported memorization baseline) already matches the headline 3D encoders; incorporating the queries (explicitly provided by the benchmark) pushes performance even higher.  

The strongest results in the entire comparison come from a simple average of the two similarity scores. That a few lines of code with no learnable parameters can outscore purpose‑built deep‑learning encoders underscores how heavily \textsc{LIT‑PCBA} rewards memorizing information disclosed by the benchmark itself, rather than modeling genuine protein–ligand recognition.

\begin{table}[ht]
\small
\setlength{\tabcolsep}{5pt}
\centering
\begin{tabular}{l
                *{4}{S[table-format=2.3]} |   
                *{4}{S[table-format=1.3]} }    
\toprule
\multirow{2}{*}{\textbf{Statistic}} &
\multicolumn{4}{c|}{\textbf{EF$_{1\%}$}} &
\multicolumn{4}{c}{\textbf{Normalized EF$_{1\%}$}} \\
\cmidrule(lr){2-5} \cmidrule(lr){6-9}
& \textbf{Qry} & \textbf{Act} & \textbf{Union} & \textbf{Avg}  &
  \textbf{Qry} & \textbf{Act} & \textbf{Union} & \textbf{Avg}  \\
\midrule
Mean   & 2.339 & 4.347 & 4.396 & \textbf{5.831}  & 0.036 & 0.063 & 0.063 & \textbf{0.076} \\
Median & 1.301 & 4.150 & 4.150 & \textbf{4.325}  & 0.013 & 0.042 & 0.042 & \textbf{0.065} \\
\bottomrule
\end{tabular}
\caption{Performance of four non-learning max-pooling strategies: \textbf{Qry} (queries only), \textbf{Act} (training actives only, equivalent to ``Mem`` in \Cref{tab:ef1_multi}), \textbf{Union} (max of query / active similarities), and \textbf{Avg} (mean of query + active similarities).  
Left block: raw EF$_{1\%}$; right block: values normalized to each target's theoretical maximum EF$_{1\%}$ (nEF$_{1\%}$).}
\label{tab:maxpool_variants}
\end{table}

\section{Why Do We Remove Stereochemistry In Our Audit?}
Because this paper centers on data leakage, we first define the term to avoid ambiguity.

\textbf{Data leakage.} We define \emph{data leakage} as unintended overlap between training and test information that lets a model achieve inflated performance by exploiting redundancy rather than learning generalizable patterns — often evidenced when trivial, untrained methods perform well by memorization.

For clarity, by \emph{label-discordant stereoisomers} we mean a stereoisomeric group in which at least one stereoisomer is labeled \emph{active} and another is labeled \emph{inactive}; this is the strongest possible supervision signal for a model to learn that stereochemistry matters.

We remove stereochemistry in this audit to \emph{expose and control} leakage, not to claim performance gains from discarding a chemically meaningful feature. In \textsc{LIT-PCBA}, stereochemical signal is essentially absent from the training data: 
\begin{itemize}
    \item 12/15 receptors have no label-discordant stereoisomers;
    \item The remaining three contain only 10 such cases; 
    \item Across the whole dataset, of 382{,}856 unique molecules, just 3{,}582 (0.94\%) have stereoisomers present, and just 202 such groups (0.053\%) exhibit conflicting activity labels.
\end{itemize}

Furthermore, stereochemistry is scarcely rewarded at evaluation: there is exactly one instance of label-discordant stereoisomers in the validation set. Meanwhile, stereoisomers recur across splits almost exclusively with the \emph{same} label. This makes stereo‑agnostic memorization appear to generalize and, in some cases, even leads to inconsistent penalties (e.g., 37 cross‑split active/inactive divergences alongside 5{,}357 duplicate inactives). Collapsing stereoisomers exposes this leakage by eliminating trivial stereo‑agnostic duplicates that can be mistaken for genuine generalization. 

While models pretrained on richer chemistry may be less affected, \textsc{LIT‑PCBA} itself provides neither a consistent training signal nor a meaningful evaluation reward for stereochemical reasoning.

These properties produce three practical consequences, detailed below.

\begin{enumerate}
    \item \textbf{No learnable stereochemical signal in training.} The few discordant examples are too sparse — 0.053\% of all molecules — to support stereochemical reasoning. 
    \item \textbf{Negligible evaluation reward.} A model that distinguishes stereoisomers gains no measurable advantage.
    \item \textbf{Label-discordant stereoisomers present but unrewarded.} There are 29 cases where a validation active has a training inactive stereoisomer, and 8 in the reverse direction (37 total), but these are overwhelmed by thousands of stereo‑agnostic duplicates. As a result, models trained on \textsc{LIT‑PCBA} will reasonably treat stereoisomers as identical, and may even be penalized for doing so.
\end{enumerate}

We fully acknowledge that stereochemistry is a chemically essential and highly informative aspect of ligand structure. Models can, in principle, benefit from stereospecific features even in the absence of many explicit stereoisomer groups with discordant labels. Our decision to remove stereochemistry in the audit is therefore methodological, not prescriptive: it isolates a specific leakage mode and clarifies what the benchmark actually measures. Many other leakage modes we identify -- such as analog leakage, high 2D‑fingerprint similarity, and query-set duplication -- are unrelated to stereochemistry and persist even after removal. This is not a recommendation for production virtual screening or medicinal chemistry workflows; the remaining leakage pathways are analyzed in subsequent sections.

\begin{table}[H]
\setlength{\tabcolsep}{1.1pt}
\renewcommand{\arraystretch}{1.06}
\centering
\begin{tabular}{
  l
  S[table-format=6.0,table-number-alignment=center]@{\hspace{20pt}}   
  S[table-format=6.0,table-number-alignment=center]@{\hspace{12pt}}  
  S[table-format=5.0,table-number-alignment=center]                   
  S[table-format=1.3,table-number-alignment=center]                   
  S[table-format=1.0,table-number-alignment=center]                   
  S[table-format=1.3,table-number-alignment=center]                   
  S[table-format=1.3,table-number-alignment=center]                   
}
\toprule
\textbf{Receptor} &
\textbf{Train} &
\textbf{NoSt} &
\textbf{St(n)} &
\textbf{St(\%)} &
\textbf{Disc.(n)} &
\textbf{Disc./St (\%)} &
\textbf{Disc./Tot (\%)} \\
\midrule
ADRB2     & 233970 & 232892 & 901  & 0.387 & 0 & 0.000 & 0.000 \\
ALDH1     & 80597  & 80101  & 431  & 0.538 & 7 & 1.624 & 0.009 \\
ESR1 (ago) & 3480   & 3441   & 35   & 1.017 & 0 & 0.000 & 0.000 \\
ESR1 (ant) & 3089   & 3058   & 27   & 0.883 & 0 & 0.000 & 0.000 \\
FEN1      & 264040 & 262705 & 1099 & 0.418 & 0 & 0.000 & 0.000 \\
GBA       & 219164 & 217622 & 1307 & 0.601 & 1 & 0.077 & 0.000 \\
IDH1      & 269694 & 267918 & 1313 & 0.490 & 0 & 0.000 & 0.000 \\
KAT2A     & 258213 & 256628 & 1277 & 0.498 & 0 & 0.000 & 0.000 \\
MAPK1     & 46548  & 46152  & 343  & 0.743 & 0 & 0.000 & 0.000 \\
MTORC1    & 24802  & 24802  & 0    & 0.000 & 0 & 0.000 & 0.000 \\
OPRK1     & 202184 & 201612 & 538  & 0.267 & 0 & 0.000 & 0.000 \\
PKM2      & 184082 & 183475 & 562  & 0.306 & 0 & 0.000 & 0.000 \\
PPARG     & 3228   & 3189   & 36   & 1.129 & 0 & 0.000 & 0.000 \\
TP53      & 2598   & 2565   & 30   & 1.170 & 0 & 0.000 & 0.000 \\
VDR       & 198047 & 196822 & 989  & 0.502 & 2 & 0.202 & 0.001 \\
\bottomrule
\end{tabular}
\caption{Stereochemical redundancy per receptor in \textsc{LIT-PCBA} training sets. Columns: \emph{Train} (total); \emph{NoSt} (no stereoisomers); \emph{St (n)} ($\geq$2 stereoisomers); \emph{St (\%)}; \emph{Disc. (n)} (label-discordant stereoisomers); \emph{Disc./St (\%)} (relative to \emph{St (n)}); \emph{Disc./Tot (\%)} (relative to \emph{Train}).}
\label{tab:stereo_ambiguity}
\end{table}

\section{Data Integrity Failures in \textsc{LIT-PCBA}}
A stereo‐agnostic memorization baseline using max-pooled ECFP4 similarity to the training actives matches the headline results of state-of-the-art 3D encoders on \textsc{LIT-PCBA}. Such performance from a model with \emph{no learning, docking, or physics} immediately raises the question: \emph{what hidden shortcuts is the dataset offering?} The remainder of this section answers that question by auditing every split for leakage and redundancy — 2D-identity and analog overlap across training, validation, and query sets, as well as stereochemical artifacts — so that the root causes of inflated scores become transparent.

Systematic inspection of the 15 targets in \textsc{LIT-PCBA} revealed widespread data leakage and redundancy. Our findings are summarized in \Cref{tab:data-integrity-summary}.

Here and throughout this section, ``2D‑identical'' refers to stereo‑agnostic identity:
two molecules are considered 2D-identical if their RDKit‑generated canonical SMILES match
\emph{after} removal of stereochemistry.

Our audit reveals four categories of data integrity failure: 

\begin{enumerate}
    \item \textbf{Inter-set 2D-identity leakage}: stereo-agnostic identical molecules are shared across data splits (e.g., identical ligands up to stereochemistry appear in both the query set and the training set).
    \item \textbf{Inter-set analog leakage}: structurally similar molecules (ECFP4 Tc $\geq 0.6$) are shared across data splits.
    \item \textbf{Intra-set 2D-identity redundancy}: stereo-agnostic identical molecules are repeated within the same data split (e.g., repetition of 2D-identical query ligands).
    \item \textbf{Intra-set analog redundancy}: structurally similar, near-duplicate molecules occur in the same data split. For example, in MTORC1, the query ligands \texttt{RAP}, \texttt{RAD}, and \texttt{ARD} exhibit extreme similarity, with pairwise Tanimoto coefficients (4096-bit ECFP4) as high as 0.93 and maximum common substructure (MCS) similarity ratios up to 0.95.

\end{enumerate}

These issues have been widely recognized as critical pitfalls in benchmark design~\cite{scior2012recognizing, lagarde2015benchmarking, xia2015benchmarking}.

The prevalence of data leakage and redundancy across query, training, and validation sets introduces significant dependencies that compromise \textsc{LIT-PCBA}’s ability to accurately assess model generalization. These dependencies lead to overestimated model performance by rewarding scaffold memorization without the need for learning. Together, these pitfalls severely undermine the dataset's credibility to function as a benchmark at all.

\begin{table}[htbp]
\centering
\begin{tabular}{@{}p{6cm}p{9cm}@{}}
\toprule
\textbf{Category} & \textbf{Description} \\
\midrule
\textbf{Inter-set 2D-identity Leakage} \newline \newline (RDKit canonical SMILES match) &
\begin{itemize}
    \item 2 unique actives shared between query and training sets;
    \item 1 active shared between query and validation sets;
    \item 2,491 unique inactives shared between training and validation sets.
\end{itemize} \\
\midrule
\textbf{Inter-set Analog Leakage} \newline \newline (4096-bit ECFP4 Tc $\geq 0.6$) &
\begin{itemize}
    \item Over 350 active-labeled training–validation analog pairs across all targets. More specifically:
    \item 323 such pairs in \textsc{ALDH1};
    \item 12 such pairs in \textsc{GBA}
    \item 8 such pairs in \textsc{MAPK1};
    \item At least one such pair in each of \textsc{FEN1}, \textsc{PKM2}, and \textsc{IDH1}.
\end{itemize} \\
\midrule
\textbf{Intra-set 2D-identity Redundancy} \newline \newline (RDKit canonical SMILES match) &
\begin{itemize}
    \item 5 unique ligands duplicated within query sets (same ligand used in multiple PDB structures);
    \item 1 active duplicated within a validation set;
    \item 15 unique actives duplicated within training sets;
    \item 789 unique inactives duplicated within validation sets;
    \item 2,945 unique inactives duplicated within training sets.
\end{itemize} \\
\midrule
\textbf{Intra-set Analog Redundancy} \newline \newline (4096-bit ECFP4 Tc $\geq 0.85$ \newline or MCS similarity $\geq 0.9$) &
\begin{itemize}
    \item In \textsc{MTORC1}, 3 of 5 query ligands have pairwise Tc up to 0.93 and MCS similarity up to 0.95;
    \item In \textsc{KAT2A}, 2 of 3 queries have Tc 0.93;
    \item In \textsc{GBA}, 2 of 5 queries have Tc 0.87.
\end{itemize} \\
\bottomrule
\end{tabular}
\caption{
Data integrity failures identified across the 15 targets in \textsc{LIT-PCBA}. 
The table is organized by four categories of data integrity failure: inter-set 2D-identity leakage (stereo-agnostic identical compounds shared between data splits), inter-set analog leakage (structurally similar compounds shared between data splits), intra-set 2D-identity redundancy (stereo-agnostic identical compounds repeated in the same data split), and intra-set analog redundancy (structurally similar compounds occurring in the same data split).
2D-identity status was defined as matching canonical SMILES generated by RDKit after removing stereochemistry. Analog criteria differed by context — inter-set analogs use a conservative threshold of 4096-bit ECFP4 Tanimoto coefficient (Tc) $\geq 0.6$ to flag scaffold-level overlap, while intra-set analogs use 4096-bit ECFP4 Tc $\geq 0.85$ or maximum common substructure (MCS) $\geq 0.9$ to capture near-duplicate redundancy. Importantly, both analog analyses were performed after removing duplicates.
}\label{tab:data-integrity-summary}
\end{table}

\subsection{Data Leakage}

\subsubsection{2D-identity Leakage Among Query, Training, and Validation Sets}
The most severe flaw is the repetition of 2D-identical molecules across data splits intended to be disjoint. Two examples: the \textsc{PKM2} query ligand \texttt{DZG} (from PDB code \texttt{3GQY}) appears in the training set, and the query ligand \texttt{D8G} (from PDB code \texttt{3H6O}) appears in the validation set. Such overlaps enable models to trivially succeed by memorizing 2D molecular identity rather than learning transferable patterns.

Furthermore, our analysis uncovered a total of 2{,}491 unique inactives shared between training and validation sets across the 15 targets. Even a single such instance compromises the assumption of independence between splits and can significantly inflate performance metrics, a known issue that has plagued earlier benchmarks like \textsc{DUD-E}~\cite{chen2019hidden, mysinger2012directory}.

\subsubsection{Analog Leakage Between Training and Validation Sets}

Beyond duplication, we observed substantial analog leakage, finding highly similar compounds between training and validation set actives. This undermines the goal of generalization to novel chemotypes.  

In real-world screening campaigns, virtual hits with Tanimoto coefficient (Tc) $> 0.35$ to known ligands are often filtered out to prioritize novel scaffolds~\cite{lyu2024alphafold2}. For our leakage analysis, we used a more conservative threshold of Tc $\geq 0.6$ to identify structurally similar analogs. 

Analog leakage is especially severe for \textsc{ALDH1}, where we identified 323 active training–validation pairs with 4096-bit ECFP4 Tc $\geq 0.6$, with many exceeding 0.9. Additional examples include \textsc{GBA} (12 pairs) and \textsc{MAPK1} (8 pairs). This type of scaffold overlap between training and testing folds is a well-documented source of bias that allows models to appear to perform well without achieving generalization~\cite{rohrer2009maximum, janet2023latentbias, kanakala2023latent}. Across all targets, we identified over 350 such analog pairs.

\subsection{Data Redundancy}

The benchmark suffers from poor diversity due to the presence of duplicate ligands and near-duplicate analogs in the query set.

\subsubsection{2D-identity Redundancy in Query Set}

For several targets, the query set contains duplicates caused by the same ligand occurring in multiple PDB co-crystal structures. In \textsc{MTORC1}, only five unique ligands are used across 11 PDB entries: \texttt{RAP} appears in five entries, \texttt{RAD} two entries, and \texttt{ARD} two entries. Thus, nine of the 11 queries (82\%) are composed of just three highly similar scaffolds. Similar cases occur in targets \textsc{VDR} (\texttt{TEJ} in two entries) and \textsc{ADRB2} (\texttt{POG} in three entries).

\subsubsection{Analog Redundancy in Query Set}

Even after dropping duplicates by canonical SMILES, there remains the issue of tightly over-represented chemotypes among query ligands. For example, in \textsc{MTORC1}, the query ligands (\texttt{RAP}, \texttt{RAD}, \texttt{ARD}) are highly similar, with pairwise Tanimoto coefficients up to 0.93 (computed using ECFP4 fingerprints of 4096 bits) and maximum common substructure (MCS) similarity ratios up to 0.95. In \textsc{KAT2A}, two of the three queries reach 0.93 similarity; in \textsc{GBA}, two of five reach 0.87 (both using the same fingerprint setup). These near duplicates dominate the query set, reducing scaffold diversity and biasing performance metrics toward models that succeed on narrow chemical classes, a long-recognized challenge in constructing useful benchmark sets~\cite{bauer2013evaluation, irwin2008community}.

\section{Implications for Virtual Screening Benchmarks}
The implications of our findings differ depending on how the \textsc{LIT-PCBA} dataset is used -- whether as a training resource for machine learning models or as a benchmarking suite for evaluating them on the validation set alone.

\subsection{Consequences for Model Training on \textsc{LIT-PCBA}}
Some issues we identify apply specifically to workflows that use the \textsc{LIT-PCBA} training data to train new models. If a model is strictly pretrained elsewhere and has had no access -- direct or indirect -- to the training actives, then these particular concerns do not apply. However, for any model trained on the \textsc{LIT-PCBA} training set, the presence of analog leakage, 2D-identity overlap, and other artifacts can lead to inflated performance that does not generalize.

\subsubsection{Analog Leakage Compromises Generalization}
For targets like \textsc{ALDH1}, where hundreds of training–validation analogs exist, models are rewarded for recognizing minor scaffold variants rather than discovering novel actives. This significantly undermines the benchmark’s goal of measuring model generalization, a long-standing challenge in virtual screening~\cite{schneider2010virtual, janet2023latentbias}. Multi-ligand models that use actives from the training set at inference run are especially vulnerable to this deficit, as training set actives directly incorporated as queries may be duplicated or nearly duplicated in the validation set, further inflating apparent performance.

\subsubsection{Label Imbalance Among Stereoisomers Misleads Learning}
From the entire training set, there are only 10 examples of label discordance, where a model can learn the importance of differentiating between stereoisomers for the purpose of identifying actives from inactives. A model trained on \textsc{LIT-PCBA} learns from 2,948 examples of same-label stereoisomers. In effect, it is taught to treat all stereoisomers as identical.  

\subsection{Impacts on Benchmark Evaluation Using Validation Data}
These issues also affect any method evaluated on the validation set, regardless of whether it was trained on \textsc{LIT-PCBA} data.

\subsubsection{2D-identity Leakage Inflates Evaluation Metrics}
In the case of \textsc{PKM2}, where the ligand \texttt{D8G} appears in both the query and validation sets, a ligand-based model using SMILES or fingerprints will assign it maximum similarity. The validation set contains 136 actives and 61{,}007 inactives ($N = 61{,}143$), so the top 0.1\% corresponds to $k = \lfloor 0.001 \times 61{,}143 \rfloor = 61$ molecules. A random model yields $\frac{136}{61{,}143} \times 61 \approx 0.136$ actives in the top 0.1\%. But with one guaranteed match, we expect to retrieve $1 + \frac{135}{61{,}142} \times 60 \approx 1.132$ actives, yielding an inflated $\mathrm{EF}_{0.1\%} = \frac{1.132}{0.136} \approx 8.32$, despite otherwise random performance.

\textbf{Remark}. We remark that this ligand in validation set is exact one (stereospecific) and not the result of us stripping stereochemistry. So this leakage happens regardless of stereochemistry. 

\subsubsection{Redundancy Biases Evaluation Metrics}
In \textsc{MTORC1}, 9 of 11 queries are either \texttt{RAP} or its nearly identical analogs. This means that performance metrics such as average AUROC or $\mathrm{EF}_{0.1\%}$ solely depends on this one chemotype and would not capture whether a model can generalize across diverse chemotypes. Even after removing duplicates of \texttt{RAP} in the query set, high Tc and MCS similarities persist. This highlights how low query diversity can lead to misleading conclusions about a model's broader utility ~\cite{lagarde2015benchmarking}.

\section{Why We Do Not Provide a Corrected \textsc{LIT-PCBA}}

We initially intended to release a cleaned version of the entire \textsc{LIT-PCBA} benchmark to address the flaws identified in this work. However, the extent and nature of the problems make such a correction infeasible.

While issues like exact duplicate removal and leakage across splits can be addressed through systematic filtering, the more fundamental flaws — such as pervasive analog leakage between training and validation sets and the low chemical diversity of query structures — cannot be resolved through simple de-duplication. These are not isolated artifacts but reflect systemic design limitations that undermine the data integrity of the benchmark to the point of irreparability.

Correcting the benchmark would require discarding large portions of the dataset, redefining the logic of data splitting, and possibly the curation of new actives and decoys altogether. Such extensive intervention would be tantamount to constructing a new benchmark from scratch rather than a faithful revision of \textsc{LIT-PCBA}.

For these reasons, we do not provide a corrected version. Instead, we recommend the construction of new benchmarks for virtual screening that use stricter controls on molecular overlap, scaffold similarity, and query diversity, echoing calls from previous benchmarking studies~\cite{xia2015benchmarking, lagarde2015benchmarking}. 

\section{Code and Data Availability}
All code, data splits, and Jupyter notebooks necessary to reproduce every number in this paper are publicly available at \href{https://github.com/sievestack/LIT-PCBA-audit}{https://github.com/sievestack/LIT-PCBA-audit}.  

\section{Appendix: Corrected \textsc{LIT-PCBA} Query Ligand SMILES}\label{sec:appendix}

The original \textsc{LIT-PCBA} query ligands are distributed as \texttt{.mol2} files that often lack bond orders, impeding accurate structure reconstruction and yielding incorrect SMILES. This affects both ligand-based methods (2D fingerprints) and structure-based workflows (e.g., docking), where bond orders define chemistry and flexibility.

We addressed this by (i) mapping each query ligand to its PDB ligand code, (ii) retrieving the corresponding structures from RCSB PDB, and (iii) generating canonical SMILES with RDKit~\cite{rdkit}. \emph{In this appendix, we retain stereochemistry} to unambiguously identify the crystallographic query ligands. \emph{In the main audit, we remove stereochemistry} (i.e., use stereo-agnostic canonical SMILES) to reveal leakage and redundancy without conflating results with stereochemical variants. The table below reports the stereo-preserved identifiers; the corresponding stereo-agnostic SMILES used in our analyses are provided with the code and data release.

\begin{footnotesize}
\begin{longtable}{llp{14cm}}
\caption{PDB codes, corresponding ligands, and canonical stereo-specific SMILES used to reference each query ligand.}\label{tab:receptor_pdb_ligands} \\
\toprule
\textbf{PDB} & \textbf{Ligand} & \textbf{SMILES} \\
\midrule
\endfirsthead

\multicolumn{3}{l}{\textit{(continued from previous page)}} \\
\toprule
\textbf{PDB} & \textbf{Ligand} & \textbf{SMILES} \\
\midrule
\endhead

\midrule
\multicolumn{3}{r}{\textit{(continued on next page)}} \\
\endfoot

\bottomrule
\endlastfoot

\multicolumn{3}{l}{\textbf{ADRB2}} \\
3P0G & P0G & \seqsplit{Cc1ccccc1CC(C)(C)NC[C@H](O)c1ccc(O)c2c1OCC(=O)N2} \\
3PDS & ERC & \seqsplit{COc1cc(CCNC[C@H](O)c2ccc(O)c3[nH]c(=O)ccc23)ccc1OCCCS} \\
3SN6 & P0G & \seqsplit{Cc1ccccc1CC(C)(C)NC[C@H](O)c1ccc(O)c2c1OCC(=O)N2} \\
4LDE & P0G & \seqsplit{Cc1ccccc1CC(C)(C)NC[C@H](O)c1ccc(O)c2c1OCC(=O)N2} \\
4LDL & XQC & \seqsplit{CC(C)(Cc1ccc(O)cc1)NC[C@H](O)c1ccc(O)c(O)c1} \\
4LDO & ALE & \seqsplit{CNC[C@H](O)c1ccc(O)c(O)c1} \\
4QKX & 35V & \seqsplit{COc1cc(CCNC[C@H](O)c2ccc(O)c(O)c2)ccc1OCCS} \\
6MXT & K5Y & \seqsplit{OCc1cc([C@@H](O)CNCCCCCCOCCCCc2ccccc2)ccc1O} \\
\\
\multicolumn{3}{l}{\textbf{ALDH1}} \\
4WP7 & 3SR & \seqsplit{CC(C)CCn1c(CN2CCN(C(=O)c3ccco3)CC2)nc2c1c(=O)n(C)c(=O)n2C} \\
4WPN & 3ST & \seqsplit{CC(C)CCn1c(CN2CCC(C(N)=O)CC2)nc2c1c(=O)n(C)c(=O)n2C} \\
4X4L & 3XG & \seqsplit{CCOC(=O)CSc1nc2c(sc3ccccc32)c(=O)n1CCCN1CCCC1} \\
5AC2 & K9P & \seqsplit{CCCCCC(=O)N1C[C@@H](C)c2c1cc(O)c1ccccc21} \\
5L2M & 6ZY & \seqsplit{Cc1oc2cc3oc(=O)c(CCC(=O)N4CCCCC4)c(C)c3cc2c1C} \\
5L2N & 6ZU & \seqsplit{Cc1c(Cc2ccccc2)c(=O)oc2cc(OS(C)(=O)=O)ccc12} \\
5L2O & 6ZW & \seqsplit{CCN(CC)c1ccc2c(C)cc(=O)oc2c1} \\
5TEI & M39 & \seqsplit{Cc1ccccc1-n1c(SCc2cccc(F)c2)nc2n[nH]cc2c1=O} \\
\\
\multicolumn{3}{l}{\textbf{ESR1 (ago)}} \\
1L2I & ETC & \seqsplit{CC[C@@H]1Cc2cc(O)ccc2C2=C1c1ccc(O)cc1C[C@H]2CC} \\
2B1V & 458 & \seqsplit{CC1=CC[C@]2(CO)CO[C@H](c3ccc(O)cc3)[C@H]1C2} \\
2B1Z & 17M & \seqsplit{C[C@]12CCc3c(ccc4cc(O)ccc34)[C@@H]1CC[C@@]2(C)O} \\
2P15 & EZT & \seqsplit{C[C@]12CC[C@@H]3c4ccc(O)cc4CC[C@H]3[C@@H]1CC[C@@]2(O)/C=C/c1ccccc1C(F)(F)F} \\
2Q70 & DC8 & \seqsplit{Oc1ccc([C@@H]2Oc3ccc(O)cc3[C@@H]3CC(F)(F)C[C@@H]32)cc1} \\
2QR9 & HZ3 & \seqsplit{COC(=O)C1=C(C(=O)OC)[C@@H]2O[C@H]1C(c1ccc(O)cc1)=C2c1ccc(O)cc1} \\
2QSE & 1HP & \seqsplit{Cn1c(N)nc2ncc(-c3ccc(O)cc3)cc21} \\
2QZO & KN1 & \seqsplit{C=CCn1nc(-c2ccc(O)cc2O)c2cccc(C(F)(F)F)c21} \\
4IVW & 1GJ & \seqsplit{Oc1ccc(-c2c3cccc(C(F)(F)F)c3nn2Cc2ccccc2)c(O)c1} \\
4PPS & ESE & \seqsplit{C[C@]12CC[C@@H](c3ccc(O)cc3)C[C@H]1CC[C@@H]2O} \\
5DRJ & 5EU & \seqsplit{Cc1cc(-c2ccc(O)cc2Cl)sc1-c1ccc(O)cc1Cl} \\
5DU5 & 5G2 & \seqsplit{O=S1(=O)C=C(c2ccc(O)cc2Cl)C(c2ccc(O)cc2Cl)=C1} \\
5DUE & 5FY & \seqsplit{Cc1cc(O)ccc1C1=C(c2ccc(O)cc2C)[C@H]2[C@@H](S(=O)(=O)Oc3ccc(O)cc3)C[C@@H]1[S@@]2=O} \\
5DZI & 5KF & \seqsplit{OCC[C@H]1CCCC(=C(c2ccc(O)cc2)c2ccc(O)cc2)C1} \\
5E1C & 5K8 & \seqsplit{COC(=O)C(C(=O)OC)[C@H]1CCCC(=C(c2ccc(O)cc2)c2ccc(O)cc2)C1} \\
\\
\multicolumn{3}{l}{\textbf{ESR1 (ant)}} \\
1XP1 & AIH & \seqsplit{C[C@H]1CN(CCOc2ccc([C@@H]3Oc4ccc(O)cc4S[C@@H]3c3ccc(O)cc3)cc2)C[C@@H]1C} \\
1XQC & AEJ & \seqsplit{Oc1ccc2c(c1)CCN(c1ccccc1)[C@@H]2c1ccc(N2CCN3CCCC[C@H]3C2)cc1} \\
2AYR & L4G & \seqsplit{CS(=O)(=O)c1ccc(-c2ccc3cc(O)ccc3c2Oc2ccc(OCCN3CCCCC3)cc2)cc1} \\
2IOG & IOG & \seqsplit{C[C@H](CCc1ccc(O)cc1)NC(=O)Cc1c(-c2ccccc2)[nH]c2cc(OCCN3CCCCC3)ccc12} \\
2IOK & IOK & \seqsplit{C[C@H](CCc1ccc(O)cc1)NC(=O)Cc1c(-c2ccccc2)[nH]c2ccccc12} \\
2OUZ & C3D & \seqsplit{Oc1ccc2c(c1)CC[C@H](c1ccccc1)[C@@H]2c1ccc(OCCN2CCCC2)cc1} \\
2POG & WST & \seqsplit{Oc1ccc([C@@H]2Oc3cccc(O)c3[C@@H]3CCC[C@@H]32)cc1} \\
2R6W & LLB & \seqsplit{CC1CCN(CCOc2ccc(C(=O)c3c(-c4ccc(O)cc4)sc4cc(O)ccc34)cc2)CC1} \\
3DT3 & 369 & \seqsplit{Cc1cc2cc(O)ccc2c(Oc2ccc(O)cc2)c1-c1cccc(O)c1} \\
5AAU & XBR & \seqsplit{O=C(O)CCN1CCc2c([nH]c3ccccc23)[C@H]1c1ccc(Cl)cc1} \\
5FQV & VQI & \seqsplit{Cc1cc2c(cc1O)CCN(CC(C)C)[C@@H]2c1ccc(/C=C/C(=O)O)cc1} \\
5T92 & 77W & \seqsplit{C[C@@]1(c2ccc(/C=C/C(=O)O)cc2)c2ccc(O)cc2CCN1c1ccc(F)cc1} \\
5UFX & 86Y & \seqsplit{CC1=C(c2ccc(O)cc2)[C@H](c2ccc(OCCN3CC[C@@H](C)C3)cc2)Oc2cc(O)ccc21} \\
6B0F & C6V & \seqsplit{CC(F)(F)c1cc(F)ccc1-c1sc2cc(O)ccc2c1Oc1ccc(/C=C/C(=O)O)cc1} \\
6CHW & F3D & \seqsplit{CC/C(=C(/c1ccc(OCCNCCCC(=O)N(C)C)cc1)c1ccc2[nH]ncc2c1)c1ccccc1} \\
\\
\multicolumn{3}{l}{\textbf{FEN1}} \\
5FV7 & R3Z & \seqsplit{O=c1c2sccc2n(C[C@H]2COc3ccccc3O2)c(=O)n1O} \\
\\
\multicolumn{3}{l}{\textbf{GBA}} \\
2V3D & NBV & \seqsplit{CCCCN1C[C@H](O)[C@@H](O)[C@H](O)[C@H]1CO} \\
2V3E & NND & \seqsplit{CCCCCCCCCN1C[C@H](O)[C@@H](O)[C@H](O)[C@H]1CO} \\
2XWD & LGS & \seqsplit{CCCCCCCC/N=C1\textbackslash\{\}OC[C@@H]2[C@@H](O)[C@H](O)[C@@H](O)[C@H](O)N12} \\
2XWE & AMF & \seqsplit{CCCCCCCC/N=C1\textbackslash\{\}SC[C@@H]2[C@@H](O)[C@H](O)[C@@H](O)[C@H](O)N12} \\
3RIK & 3RI & \seqsplit{OCCN1C[C@H](O)[C@@H](O)[C@H](O)[C@@H](O)C1} \\
3RIL & 3RK & \seqsplit{O[C@H]1[C@H](O)[C@@H](O)CNC[C@@H]1O} \\
\\
\multicolumn{3}{l}{\textbf{IDH1}} \\
4I3K & 1BX & \seqsplit{Cc1cc(Cc2ccc(O)cc2)n(O)c(=O)c1} \\
4I3L & 1BZ & \seqsplit{Cc1cc(Cc2ccccc2)n(O)c(=O)c1} \\
4UMX & VVS & \seqsplit{CC(C)(C)CC(C)(C)c1cc(Cn2ccnc2)c(O)c(Cn2ccnc2)c1} \\
4XRX & 42V & \seqsplit{CN1C(=O)/C(=C\textbackslash\{\}c2ccc(=O)[nH]c2)NC1=S} \\
4XS3 & 42W & \seqsplit{CN1C(=O)/C(=C\textbackslash\{\}c2ccc(=O)n(Cc3ccccc3)c2)NC1=S} \\
5DE1 & 59D & \seqsplit{C[C@H](O)c1cccc(NC(=O)c2nn(Cc3ccc(F)cc3)c3c2CN(C(=O)c2ccc[nH]2)C[C@H]3C)c1} \\
5L57 & 6N3 & \seqsplit{CC(C)CCOc1nc(N2C[C@@H]3[C@H](C2)[C@H]3C(=O)O)ccc1C(=O)N[C@@H]1[C@@H]2C[C@@H]3C[C@H]1C[C@@](O)(C3)C2} \\
5L58 & 6MX & \seqsplit{O=C(O)C[C@H]1CCN(c2ccc(C(=O)N[C@@H]3[C@@H]4C[C@@H]5C[C@H]3C[C@@](O)(C5)C4)c(SC3CCCCC3)n2)C1} \\
5LGE & 6VN & \seqsplit{CC(C)c1ccc(Nc2nc3cc(C(=O)O)ccc3n2[C@H]2C[C@@H](C)CC(C)(C)C2)cc1} \\
5SUN & 70Q & \seqsplit{CN(C)S(=O)(=O)c1cccc(NC(=O)c2nn(Cc3ccccc3)c(=O)c3ccccc23)c1} \\
5SVF & 70P & \seqsplit{CC(C)[C@H]1COC(=O)N1c1ccnc(N[C@@H](C)c2ccccc2)n1} \\
5TQH & 7J2 & \seqsplit{Cc1cc(-c2cnc([C@H](C)Nc3nccc(N4C(=O)OC[C@@H]4C(C)C)n3)nc2)ccc1F} \\
6ADG & 9UO & \seqsplit{C[C@@H](Nc1nc(N[C@H](C)C(F)(F)F)nc(-c2cccc(Cl)n2)n1)C(F)(F)F} \\
6B0Z & C81 & \seqsplit{Cc1cc([C@H](C)Nc2nccc(N3C(=O)OC[C@@H]3[C@H](C)F)n2)ncc1-c1ccnc(C(F)(F)F)c1} \\
\\
\multicolumn{3}{l}{\textbf{KAT2A}} \\
5H84 & 1VU & \seqsplit{CCC(=O)SCCNC(=O)CCNC(=O)[C@H](O)C(C)(C)CO[P@](=O)(O)O[P@@](=O)(O)OC[C@H]1O[C@@H](n2cnc3c(N)ncnc32)[C@H](O)[C@@H]1OP(=O)(O)O} \\
5H86 & BCO & \seqsplit{CCCC(=O)SCCNC(=O)CCNC(=O)[C@H](O)C(C)(C)CO[P@](=O)(O)O[P@@](=O)(O)OC[C@H]1O[C@@H](n2cnc3c(N)ncnc32)[C@H](O)[C@@H]1OP(=O)(O)O} \\
5MLJ & 9ST & \seqsplit{CN1C[C@H](Nc2cnn(C)c(=O)c2Br)C[C@H](c2ccccc2)C1} \\
\\
\multicolumn{3}{l}{\textbf{MAPK1}} \\
1PME & SB2 & \seqsplit{C[S@](=O)c1ccc(-c2nc(-c3ccc(F)cc3)c(-c3ccncc3)[nH]2)cc1} \\
2OJG & 19A & \seqsplit{CN(C)C(=O)c1cc(-c2n[nH]cc2-c2ccccc2)c[nH]1} \\
3SA0 & NRA & \seqsplit{O=c1c2cc(O)c(O)cc2oc2cc(O)cc(O)c12} \\
3W55 & 1FM & \seqsplit{COc1cc(O)c2c(c1)/C=C/C[C@H](O)[C@H](O)C(=O)/C=C\textbackslash\{\}C[C@H](C)OC2=O} \\
4QP3 & 36Q & \seqsplit{OC[C@H](Cc1ccccc1)Nc1ncnc2[nH]cnc12} \\
4QP4 & 36O & \seqsplit{c1nc(NC2CCCCC2)c2nc[nH]c2n1} \\
4QP9 & 35X & \seqsplit{CCCn1cc(-c2c[nH]c3ncc(-c4ccncc4)nc23)cn1} \\
4QTA & 38Z & \seqsplit{O=C(Nc1ccc2n[nH]c(-c3ccncc3)c2c1)[C@@H]1CCN(CC(=O)N2CCN(c3ccc(-c4ncccn4)cc3)CC2)C1} \\
4QTE & 390 & \seqsplit{Cc1cnc(Nc2ccc(F)cc2Cl)nc1-c1c[nH]c(C(=O)N[C@H](CO)c2cccc(Cl)c2)c1} \\
4XJ0 & 41B & \seqsplit{O=c1cc(-c2ccnc(NC3CCOCC3)n2)ccn1[C@H](CO)c1ccc(Cl)c(F)c1} \\
4ZZN & CQ8 & \seqsplit{CNC(=O)c1ccccc1Nc1cc(NC2CCOCC2)ncc1Cl} \\
5AX3 & 5ID & \seqsplit{Nc1ncnc2c1c(I)cn2[C@@H]1O[C@H](CO)[C@@H](O)[C@H]1O} \\
5BUJ & 4VB & \seqsplit{O=c1cc(N2CCc3n[nH]c(-c4ccncc4)c3C2)cc[nH]1} \\
5V62 & FRZ & \seqsplit{Nc1n[nH]c2nnc(-c3c(-c4ccccc4)nn4ccccc34)cc12} \\
6G9H & ERW & \seqsplit{CN(C(=O)CN1Cc2ccc(-c3nc(NC4CCOCC4)ncc3Cl)cc2C1=O)C(C)(C)C} \\
\\
\multicolumn{3}{l}{\textbf{MTORC1}} \\
1FAP & RAP & \seqsplit{CO[C@H]1C[C@@H]2CC[C@@H](C)[C@@](O)(O2)C(=O)C(=O)N2CCCC[C@H]2C(=O)O[C@H]([C@H](C)C[C@@H]2CC[C@@H](O)[C@H](OC)C2)CC(=O)[C@H](C)/C=C(\textbackslash\{\}C)[C@@H](O)[C@@H](OC)C(=O)[C@H](C)C[C@H](C)/C=C/C=C/C=C/1C} \\
1NSG & RAD & \seqsplit{CCO[C@H]1C[C@@H]2CC[C@@H](C)[C@@](O)(O2)C(=O)C(=O)N2CCCC[C@H]2C(=O)O[C@H]([C@H](C)C[C@@H]2CC[C@@H](O)[C@H](OC)C2)CC(=O)[C@H](C)/C=C(\textbackslash\{\}C)[C@@H](O)[C@@H](OC)C(=O)[C@H](C)C[C@H](C)/C=C/C=C/C=C/1C} \\
2FAP & RAD & \seqsplit{CCO[C@H]1C[C@@H]2CC[C@@H](C)[C@@](O)(O2)C(=O)C(=O)N2CCCC[C@H]2C(=O)O[C@H]([C@H](C)C[C@@H]2CC[C@@H](O)[C@H](OC)C2)CC(=O)[C@H](C)/C=C(\textbackslash\{\}C)[C@@H](O)[C@@H](OC)C(=O)[C@H](C)C[C@H](C)/C=C/C=C/C=C/1C} \\
3FAP & ARD & \seqsplit{CO[C@@H]1C[C@H](C[C@@H](C)[C@@H]2CC(=O)[C@H](C)/C=C(\textbackslash\{\}C)[C@@H](O)[C@@H](OC)C(=O)[C@H](C)C[C@H](C)/C=C/C=C/C=C(\textbackslash\{\}C)[C@H](c3ccc(C)s3)C[C@@H]3CC[C@@H](C)[C@@](O)(O3)C(=O)C(=O)N3CCCC[C@H]3C(=O)O2)CC[C@H]1O} \\
4DRH & RAP & \seqsplit{CO[C@H]1C[C@@H]2CC[C@@H](C)[C@@](O)(O2)C(=O)C(=O)N2CCCC[C@H]2C(=O)O[C@H]([C@H](C)C[C@@H]2CC[C@@H](O)[C@H](OC)C2)CC(=O)[C@H](C)/C=C(\textbackslash\{\}C)[C@@H](O)[C@@H](OC)C(=O)[C@H](C)C[C@H](C)/C=C/C=C/C=C/1C} \\
4DRI & RAP & \seqsplit{CO[C@H]1C[C@@H]2CC[C@@H](C)[C@@](O)(O2)C(=O)C(=O)N2CCCC[C@H]2C(=O)O[C@H]([C@H](C)C[C@@H]2CC[C@@H](O)[C@H](OC)C2)CC(=O)[C@H](C)/C=C(\textbackslash\{\}C)[C@@H](O)[C@@H](OC)C(=O)[C@H](C)C[C@H](C)/C=C/C=C/C=C/1C} \\
4DRJ & RAP & \seqsplit{CO[C@H]1C[C@@H]2CC[C@@H](C)[C@@](O)(O2)C(=O)C(=O)N2CCCC[C@H]2C(=O)O[C@H]([C@H](C)C[C@@H]2CC[C@@H](O)[C@H](OC)C2)CC(=O)[C@H](C)/C=C(\textbackslash\{\}C)[C@@H](O)[C@@H](OC)C(=O)[C@H](C)C[C@H](C)/C=C/C=C/C=C/1C} \\
4FAP & ARD & \seqsplit{CO[C@@H]1C[C@H](C[C@@H](C)[C@@H]2CC(=O)[C@H](C)/C=C(\textbackslash\{\}C)[C@@H](O)[C@@H](OC)C(=O)[C@H](C)C[C@H](C)/C=C/C=C/C=C(\textbackslash\{\}C)[C@H](c3ccc(C)s3)C[C@@H]3CC[C@@H](C)[C@@](O)(O3)C(=O)C(=O)N3CCCC[C@H]3C(=O)O2)CC[C@H]1O} \\
4JSX & 17G & \seqsplit{Nc1ccc(-c2ccc3ncc4ccc(=O)n(-c5cccc(C(F)(F)F)c5)c4c3c2)cn1} \\
4JT5 & P2X & \seqsplit{CC(C)n1nc(-c2cc3cc(O)ccc3[nH]2)c2c(N)ncnc21} \\
5GPG & RAP & \seqsplit{CO[C@H]1C[C@@H]2CC[C@@H](C)[C@@](O)(O2)C(=O)C(=O)N2CCCC[C@H]2C(=O)O[C@H]([C@H](C)C[C@@H]2CC[C@@H](O)[C@H](OC)C2)CC(=O)[C@H](C)/C=C(\textbackslash\{\}C)[C@@H](O)[C@@H](OC)C(=O)[C@H](C)C[C@H](C)/C=C/C=C/C=C/1C} \\
\\
\multicolumn{3}{l}{\textbf{OPRK1}} \\
6B73 & CVV & \seqsplit{O=C(N[C@@H]1C=C[C@H]2[C@H]3Cc4ccc(O)c5c4[C@@]2(CCN3CC2CC2)[C@H]1O5)c1cccc(I)c1} \\
\\
\multicolumn{3}{l}{\textbf{PKM2}} \\
3GQY & DZG & \seqsplit{COc1ccc(S(=O)(=O)N2CCN(S(=O)(=O)c3ccc4c(c3)OCCO4)CC2)cc1} \\
3GR4 & DYY & \seqsplit{O=S(=O)(c1ccc2c(c1)OCCO2)N1CCN(S(=O)(=O)c2c(F)cccc2F)CC1} \\
3H6O & D8G & \seqsplit{Cc1cc2c(s1)c1cnn(Cc3ccccc3F)c(=O)c1n2C} \\
3ME3 & 3SZ & \seqsplit{Nc1cccc(S(=O)(=O)N2CCCN(S(=O)(=O)c3ccc4c(c3)OCCO4)CC2)c1} \\
3U2Z & 07T & \seqsplit{Cn1c2cc([S@@](C)=O)sc2c2cnn(Cc3cccc(N)c3)c(=O)c21} \\
4G1N & NZT & \seqsplit{O=C(c1ccc(NS(=O)(=O)c2cccc3cccnc23)cc1)N1CCN(c2cnccn2)CC1} \\
4JPG & 1OX & \seqsplit{O=c1cc(Cn2cnc3ccccc32)nc2ccccn12} \\
5X1V & 7XX & \seqsplit{Cn1cc(C(=O)c2cccc(Cl)c2Cl)cc1C(N)=O} \\
5X1W & 7Y0 & \seqsplit{Cn1cc(C(=O)c2cccc(Cl)c2Cl)cc1C(=O)NCCNC(=O)c1cc(C(=O)c2cccc(Cl)c2Cl)cn1C} \\
\\
\multicolumn{3}{l}{\textbf{PPARG}} \\
1ZGY & BRL & \seqsplit{CN(CCOc1ccc(C[C@@H]2SC(=O)NC2=O)cc1)c1ccccn1} \\
2I4J & DRJ & \seqsplit{CCCCCCCN(CCc1ccc(O[C@](C)(CC)C(=O)O)cc1)c1nc2ccccc2o1} \\
2P4Y & C03 & \seqsplit{COc1ccc2c(-c3c(C)n(Cc4cc(O[C@H](C)C(=O)O)ccc4Cl)c4cc(OC(F)(F)F)ccc34)noc2c1} \\
2Q5S & NZA & \seqsplit{O=C(O)c1c(Sc2ccccc2)c2cc(Cl)ccc2n1Cc1ccc(Cl)cc1} \\
2YFE & YFE & \seqsplit{COc1cc(CCc2ccccc2)c(C(=O)O)c(O)c1CC=C(C)C} \\
3B1M & KRC & \seqsplit{CCc1ccc2ccccc2c1CNC(=O)c1c(OC)cc(O)c2c1OC1=CC(O)=C(C(C)=O)C(=O)[C@]12C} \\
3HOD & ZZH & \seqsplit{O=C(O)[C@H](Cc1ccccc1)Oc1ccc(Cc2ccccc2)cc1} \\
3R8A & HIG & \seqsplit{CCc1nc2c(C)cc(C)nc2n1[C@H]1CCc2cc(-c3ccccc3-c3nnn[nH]3)ccc21} \\
4CI5 & Y1N & \seqsplit{CC(C)(Oc1ccc(CCOc2ccc(/N=N/c3ccccc3)cc2)cc1)C(=O)O} \\
4FGY & 0W3 & \seqsplit{C[C@H](CCC(=O)O)C[C@H](C)C[C@H](C)C(=O)CC(=O)[C@H](C)C[C@H](C)C/C=C\textbackslash\{\}[C@@H](C)[C@@H](O)[C@@H](C)[C@@H](O)C[C@@H]1CC[C@@](C)([C@H]2CC[C@@](C)([C@@H](C)O)O2)O1} \\
4PRG & 072 & \seqsplit{CCCCCCC[C@@H]1S[C@@H](CC(=O)N(Cc2ccccc2)Cc2ccccc2)C(=O)N1CCCCc1ccc(C(=O)O)cc1} \\
5TTO & 7KK & \seqsplit{O=S(=O)(Nc1cc(Cl)c(Oc2cnc3ccccc3c2)c(Cl)c1)c1cc(F)c(Br)cc1F} \\
5TWO & 7MV & \seqsplit{O=C(NCc1ccccc1)c1ccc2c(ccn2Cc2ccc(Cl)c(F)c2)c1} \\
5Y2T & 8LX & \seqsplit{COc1ccc(Oc2cc(N(C)CCOc3ccc(C[C@@H]4SC(=O)NC4=O)cc3)ncn2)cc1} \\
5Z5S & RTE & \seqsplit{Cn1c(COc2cccc(C(=O)O)c2)nc2ccc(Oc3ccc(Cl)c(F)c3)cc21} \\
\\
\multicolumn{3}{l}{\textbf{TP53}} \\
2VUK & P83 & \seqsplit{CCn1c2ccccc2c2cc(CNC)ccc21} \\
3ZME & QC5 & \seqsplit{CN(C)CCn1cc(-c2ccc(F)cc2)c(-n2cccc2)n1} \\
4AGO & P74 & \seqsplit{CCN(CC)C1CCN(Cc2cc(C\#CCNC(=O)OC(C)(C)C)cc(I)c2O)CC1} \\
4AGQ & P96 & \seqsplit{CCN(CC)C1CCN(Cc2cc(C\#CCNc3ccccc3)cc(I)c2O)CC1} \\
5G4O & O80 & \seqsplit{CN(C)Cc1ccc2c(c1)c1ccccc1n2CC(F)(F)F} \\
5O1I & 9GH & \seqsplit{CCN(CC)c1nc2c(C(=O)O)c(O)c(I)c(-n3cccc3)c2s1} \\
\\
\multicolumn{3}{l}{\textbf{VDR}} \\
3A2I & TEJ & \seqsplit{C=C1C[C@H](C[C@H](C)[C@H]2CC[C@H]3/C(=C/C=C4/C[C@@H](O)C[C@H](O)C4=C)CCC[C@]23C)OC1=O} \\
3A2J & TEJ & \seqsplit{C=C1C[C@H](C[C@H](C)[C@H]2CC[C@H]3/C(=C/C=C4/C[C@@H](O)C[C@H](O)C4=C)CCC[C@]23C)OC1=O} \\
\addlinespace

\end{longtable}
\end{footnotesize}

\bibliographystyle{unsrt}
\bibliography{lit-pcba-audit}

\end{document}